\documentclass{article}
\usepackage{amsmath, amssymb, graphicx, float}

\title{Semantic Tokens in Retrieval Augmented Generation}
\author{
 Joel Suro\\
  \texttt{jsv2858@gmail.com}
}
\date{}

\begin{document}

\maketitle

\begin{abstract} Retrieval-Augmented Generation (RAG) architectures have recently garnered significant attention for their ability to improve truth grounding and coherence in natural language processing tasks. However, the reliability of RAG systems in producing accurate answers diminishes as the volume of data they access increases. Even with smaller datasets, these systems occasionally fail to address simple queries. This issue arises from their dependence on state-of-the-art large language models (LLMs), which can introduce uncertainty into the system's outputs. In this work, I propose a novel Comparative RAG system that introduces an evaluator module to bridge the gap between probabilistic RAG systems and deterministically verifiable responses. The evaluator compares external recommendations with the retrieved document chunks, adding a decision-making layer that enhances the system's reliability. This approach ensures that the chunks retrieved are both semantically relevant and logically consistent with deterministic insights, thereby improving the accuracy and overall efficiency of RAG systems. This framework paves the way for more reliable and scalable question-answering applications in domains requiring high precision and verifiability. \end{abstract}

\section{Introduction}
Recent advancements in Retrieval-Augmented Generation (RAG) architectures have significantly enhanced truth grounding and coherence in natural language processing (NLP) \cite{2024arXiv240808921P}. However, a critical challenge persists: the reliability of RAG systems diminishes as the volume of accessible data increases, leading to inconsistent or inaccurate responses \cite{anthropic2024}. Even with smaller datasets, these systems occasionally fail to address straightforward queries, largely due to their reliance on the probabilistic outputs of state-of-the-art large language models.

This issue is particularly pressing because the scalability and trustworthiness of RAG systems are vital for their deployment in real-world applications, such as medical diagnostics, legal research, and customer support, where precision and reliability are non-negotiable. Addressing these limitations is essential to realizing the full potential of RAG systems in domains demanding high accuracy and scalability.

This work introduces a conceptual framework for a RAG system that combines probabilistic reasoning with deterministic verification mechanisms. By bridging the gap between probabilistic and verifiable response generation, this framework seeks to enhance the reliability and accuracy of RAG architectures, irrespective of dataset size. This preprint outlines the foundational principles of the proposed system, explores potential implementation pathways, and discusses its broader implications for advancing state-of-the-art NLP. Moreover, this approach may extend beyond traditional RAG systems, potentially fitting into a wider class of algorithms such as GraphRag, paving the way for new paradigms in retrieval-augmented methodologies.

To illustrate the application of the proposed algorithm, consider the following scenario:

You are the manager of a food delivery company that primarily operates through a mobile application. One key objective is to determine which restaurants should be featured prominently in the hero section of the app. To achieve this, you calculate a series of metrics to develop a "desirability index," quantifying the appeal of restaurants based on factors like customer reviews, delivery times, and order volumes. While the metrics are successfully deployed, a challenge arises: the desirability index cannot be seamlessly integrated into the existing state-of-the-art RAG system that powers user queries, such as "What are the best options for Italian food nearby?"

Since the RAG system directly influences user decisions, it is essential that recommendations not only align semantically with the user's query but also incorporate deterministic insights from the desirability index. Without this integration, the RAG system may prioritize restaurants based solely on semantic relevance, potentially overlooking high-desirability options. This limitation highlights the need for a hybrid approach that combines the semantic-probabilistic reasoning common in large language models (LLMs) with deterministic verification, ensuring the most accurate and relevant recommendations for end-users \cite{2024arXiv240809031L}.

\section{Comparative RAG}

A comparative RAG system is a flexible and adaptive approach that enhances a Retrieval-Augmented Generation (RAG) pipeline by incorporating additional modules that regulate the quality of the responses. The key innovation of this system is the introduction of an evaluator module, which serves as a bridge between the retrieved data and external reasoning processes.

In a standard RAG system, the process is divided into two main stages: retrieval and augmentation. First, the user query is passed to the retrieval component, which retrieves the 
n
n most relevant chunks of information. These retrieved chunks are then fed into a large language model (LLM), which synthesizes the retrieved information to generate a response to the user's query. The later stages of this process—augmentation and generation—are where the LLM refines the answer and produces a coherent response.

The proposed approach introduces an additional module, called the evaluator, which plays a critical role regardless of the RAG system's complexity. The evaluator’s task is to compare the recommendations made by an external system against the chunks retrieved by the retrieval module. The evaluator itself decides whether a specific chunk-object is deterministically parsed (given by the external algorithm's score) to the semantic LLM that generates the final response. The simplicity of this approach allows for virtually any type of RAG system to use this framework, ranging from basic RAG to more advanced GraphRAG architectures.

The success of this integration relies on proper preprocessing and chunking of the text. Specifically, we standardize the size of the chunks and ensure that each chunk has relevant properties. The chunking process is handled by an LLM, which is instructed to "synthesize the information into 
n
n-sized chunks." This ensures that each chunk in the referenced document is treated as an independent object, complete with its own properties. Chunk-property relevance refers to the idea that each chunk represents a collection of tokens related to a single object or a set of related properties. A property, in this case, refers to a feature or description of that object.

With these assumptions in place, the evaluator can efficiently compare the retrieved chunks against external recommendations. This comparison is not only semantically meaningful but also allows the evaluator to generate a ranked list of recommended items, grounded in both the text's content and the external system’s recommendations.

In essence, the evaluator acts as an intermediary between the knowledge encoded in the referenced text and the reasoning engine of an external system. By doing so, the comparative RAG approach introduces judgment-based reasoning capabilities to an otherwise highly efficient, though probabilistic, language model. The effectiveness of this approach is influenced by both the quality of chunking and the fine-tuning of the LLM, which ensures that the data is well-structured for comparison.

\begin{figure}[H]     \centering     \includegraphics[width=\textwidth]{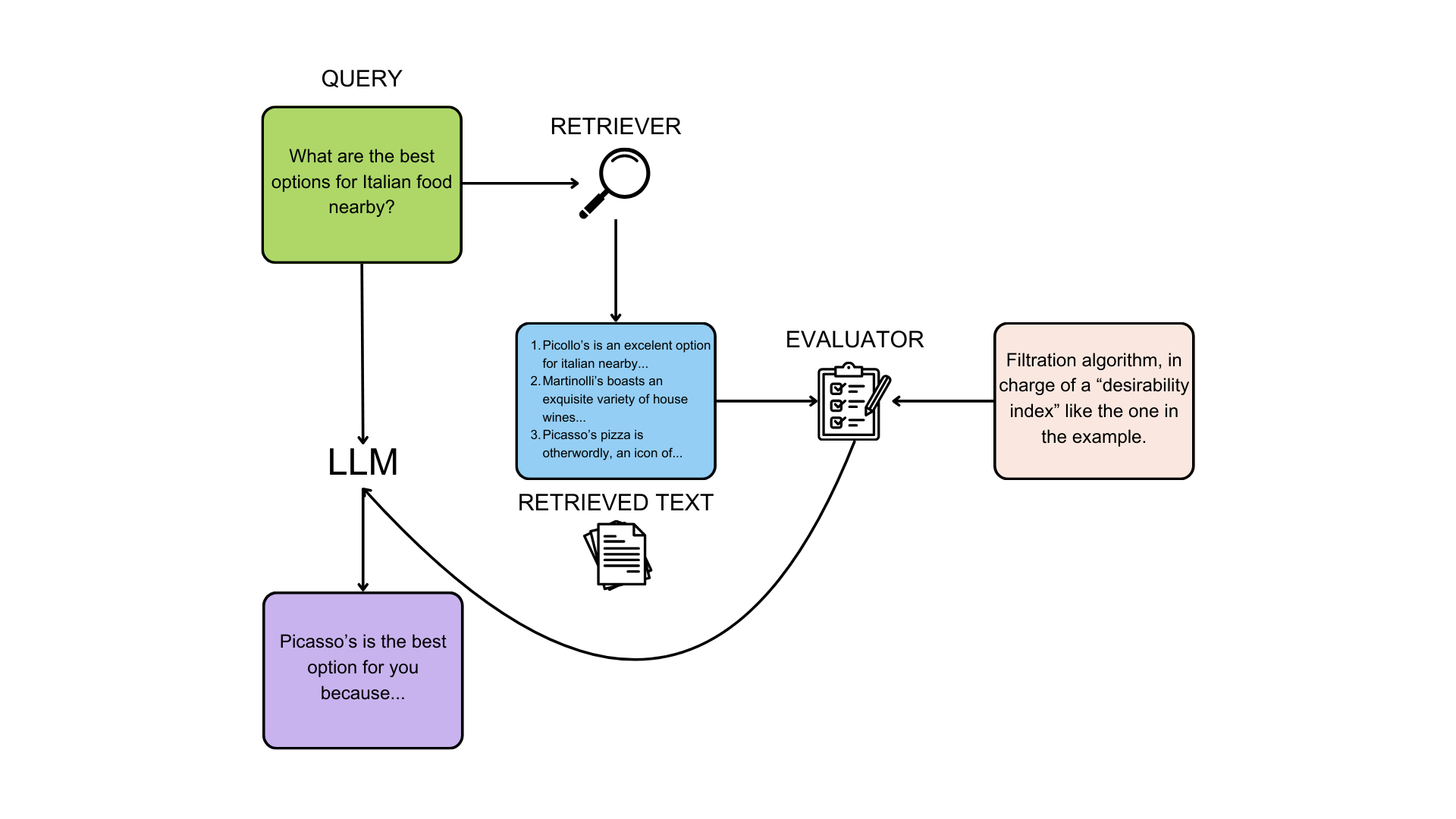}     \caption{Comparative RAG}     \label{fig:example} \end{figure}

This system can easily be expanded to more complex architectures. For example, an additional algorithm could be responsible for filtering and selecting the chunks that exhibit chunk-property relevance. Clever engineering, in this case, may allow for more sophisticated systems that integrate with this token-based RAG approach. Nevertheless, the main challenge of this model lies in the complexity and reliability of the chunk-object relevance process. As long as this assumption can be met, the system offers a consistent method of addressing some of the problems related to hallucination.

\section{Evaluator}

The \textbf{Evaluator} module enables out-of-model deterministic reasoning to interact seamlessly with semantic retrieval processes. Its primary role is to assign identifiable and unique tokens (e.g., hashes) to the headers or footers of chunk-objects. This mechanism allows for effective correlation between in-model and out-of-model data. The evaluator's tasks can be summarized as follows:

\begin{enumerate}
    \item \textbf{Preprocessing Chunks:} Analyze and organize chunks based on their relevance to the chunk-object.
    \item \textbf{Hash Assignment:} Generate and apply unique identifiers to preprocessed chunks, ensuring a direct relationship between out-of-model results and in-model chunk-objects.
\end{enumerate}

The complexity of the evaluator algorithm can range from straightforward token matching and ranking to advanced operations such as automatic preprocessing, sorting, and resolving semantic matches between chunk-objects and their out-of-model counterparts.

\subsection{Example: Food Delivery Scenario}

Consider a food delivery use case. An out-of-model ranking algorithm processes multiple metrics, including:  
\begin{itemize}
    \item NPS scores,
    \item response times,
    \item written reviews, and
    \item geographic proximity to other user-preferred activities.
\end{itemize}

These metrics are synthesized into a single \textit{desirability index} or \textit{filtration score}. Concurrently, the evaluator module applies hashes to the header or footer slots of chunk-objects, correlating them to rankings in the filtration list. This ensures a one-to-one relationship between a chunk-object and its corresponding out-of-model ranking.

For instance, the chunk-object \texttt{"Gio's"} receives a unique hash, aligning it with its corresponding filtration score in the sorted ranking array. When the RAG system processes a user query, it computes semantic results ordered by chunk-object relevance. These results are further filtered by matching the evaluator module's hash criteria with the filtration list. The outcome is a prefiltered semantic result set that aligns in-model relevance with out-of-model rankings.

\section{Conclusion}
In this work, I have proposed a novel Comparative Retrieval-Augmented Generation (RAG) system that introduces an evaluator module to enhance the standard RAG framework. The evaluator adds a decision-making layer, comparing the retrieved chunks with an external reasoning process. This approach bridges the gap between probabilistic language model outputs and more deterministic, logical decision-making, ensuring that the generated responses are not only semantically relevant but also grounded in a structured evaluation of the retrieved information.

By standardizing chunk sizes and ensuring that each chunk is relevant to a specific property or concept, the evaluator can efficiently compare the retrieved chunks with the synthesized content. This enables the generation of responses that align more closely with the user's query, while maintaining coherence and reliability. The proposed system is versatile and can function with different RAG architectures, ranging from simpler models to more complex ones.

The integration of the evaluator module underscores the potential to enhance the reliability and accuracy of RAG systems by incorporating deterministic reasoning within the probabilistic framework of large language models. This approach paves the way for more refined RAG pipelines that offer not only better semantic relevance but also improved logical consistency in the final generated responses. Moving forward, this method can be extended to more complex architectures, making RAG systems more scalable and robust in handling diverse queries. Ultimately, the comparative RAG system represents a step toward more efficient and reliable natural language processing models.

\bibliographystyle{plain}
\bibliography{references}

\cite{2024arXiv240808921P}
\cite{2024arXiv240809031L}
\cite{anthropic2024}

\end{document}